\documentclass[10pt,twocolumn,letterpaper]{article}

\usepackage{iccv}
\usepackage{times}
\usepackage{epsfig}
\usepackage{graphicx}
\usepackage{amsmath}
\usepackage{amssymb}
\usepackage{authblk}

\usepackage[pagebackref=true,breaklinks=true,letterpaper=true,colorlinks,bookmarks=false]{hyperref}

\iccvfinalcopy % *** Uncomment this line for the final submission

\def\iccvPaperID{****} % *** Enter the ICCV Paper ID here

\ificcvfinal\fi
\begin{document}

%%%%%%%%% TITLE
\title{1st Place Solutions for UG2+ Challenge 2021 - (Semi-) supervised Face detection in the low light condition}
\author{Pengcheng Wang, Lingqiao Ji, Zhilong Ji, Yuan Gao, Xiao Liu \authorcr Tomorrow Advancing Life (TAL) Education Group \authorcr 
{\tt\small \{wangpengcheng2, jilingqiao, jizhilong, gaoyuan23, liuxiao15\}@tal.com}}
\maketitle
%\thispagestyle{empty}

%%%%%%%%% ABSTRACT
% \begin{abstract}
% In this technical report, we briefly introduce the solutions of our team “TAL-ai” for (Semi-) supervised Face detection in the low light condition in UG2+Challenge in CVPR 2021. By conducting several experiments with popular image enhancement methods and image transfer methods, we observe that adapt different kinds of processed data to training can achieve better performance. We also adopt several popular object detection frameworks, e.g., DetectoRS, Cascade-RCNN, and large backbone like Swin-transformer. Finally, we ensemble several models which achieved mAP 74.89 on the testing set, ranking 1st on the final leaderboard. 
% \end{abstract}

%% wpc改过
\begin{abstract}
In this technical report, we briefly introduce the solution of our team “TAL-ai” for (Semi-) supervised Face detection in the low light condition in UG2$^{+}$ Challenge in CVPR 2021. By conducting several experiments with popular image enhancement methods and image transfer methods, we pulled the low light image and the normal image to a more closer domain. And it is observed that using these data to training can achieve better performance. We also adapt several popular object detection frameworks, e.g., DetectoRS, Cascade-RCNN, and large backbone like Swin-transformer. Finally, we ensemble several models which achieved mAP 74.89 on the testing set, ranking 1st on the final leaderboard. 
\end{abstract}
%%

%%%%%%%%% BODY TEXT
\section{Introduction}

\begin{figure}[t] 
        \centering
        \includegraphics[width=1.0 \linewidth]{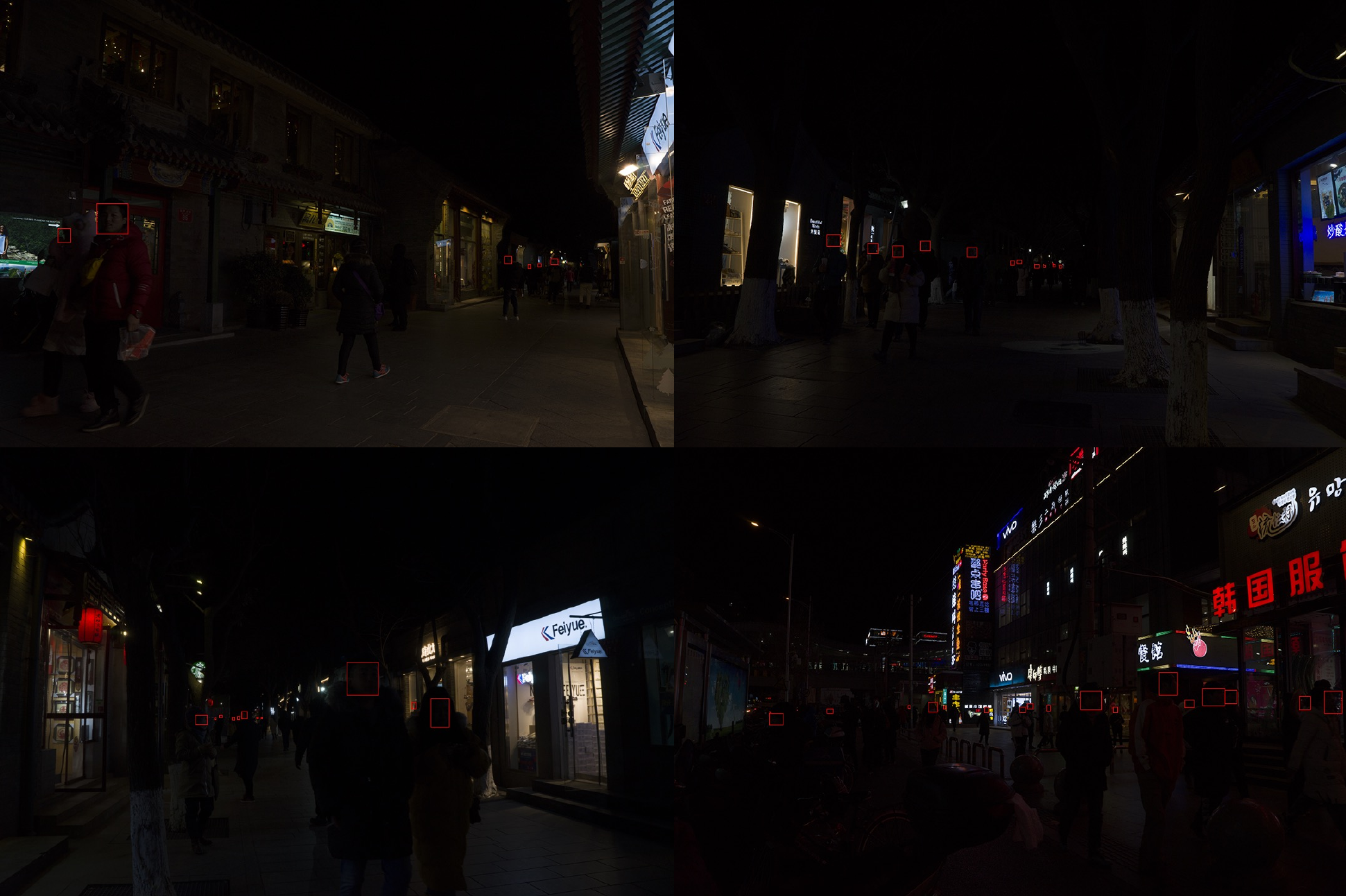}
        % \vspace{-0.3em}
        \caption{DARKFACE Dataset samples, the red boxes are ground truth.}
        \label{fig:fig1}
\vspace{-1em}
\end{figure}

The (Semi-) supervised Face detection in the low light condition challenge at CVPR 2021 is a part of the Workshop on UG2+ Prize Challenge. In this task, we need to detect faces in the low light image. Moreover, the given DARKFACE set has 6000 low light images with the corresponding face annotations (some are extremely low light conditions as shown in Figure \ref{fig:fig1}) as the training and validation sets, and the final test set consists of 4000 low light images. According to \cite{9049390}, these samples were collected on several busy streets around Beijing,  where contain faces of various scales and poses, and the resolution of these images is 1080 ×720 (down-sampled from 6K × 4K). 

%% wpc改过
\section{Overview of Methods}
In the previous work that states in \cite{9049390}, the two stages method achieves best result. Their methods usually with a WIDERFACE \cite{7780965} pre-trained model, then fine-tuned on the properly pre-processed DARKFACE set. We follow this idea to explore our methods, but differently, the training set is not just including pre-processed DARKFACE set but external sets. Additionally, we use WIDERFACE and UFDD \cite{nada2018pushing} as our external sets.

In this work, we have experimented with both image enhancement methods \cite{597272,9157813} and several object detection frameworks. For the image enhancement methods, we follow the experimental setting in \cite{597272} and \cite{9157813} to process the given low light images. Besides, following \cite{wang2021hlaface}, we transfer the normal images like WIDERFACE, UFDD dataset to the closer domain of the processed DARKFACE images. Moreover, we aggregate the saliency map \cite{5963689} of each image to the input of the network for suppressing the false negative result. After that, we evaluate the performance of the different object detection frameworks \cite{8917599,qiao2020detectors,dai2017deformable}. All the experiment results and conclusions are subsequently given.

%%%%%%%%%%%%%% figure 1 %%%%%%%%%%%%
%%%%%%%%%%%%%%%%%%%%%%

%-------------------------------------------------------------------------

%%%%%%%%%%%%% figure 2 %%%%%%%%%%%%
\begin{figure*}[h] 
        \centering
        \includegraphics[width=1.0 \linewidth]{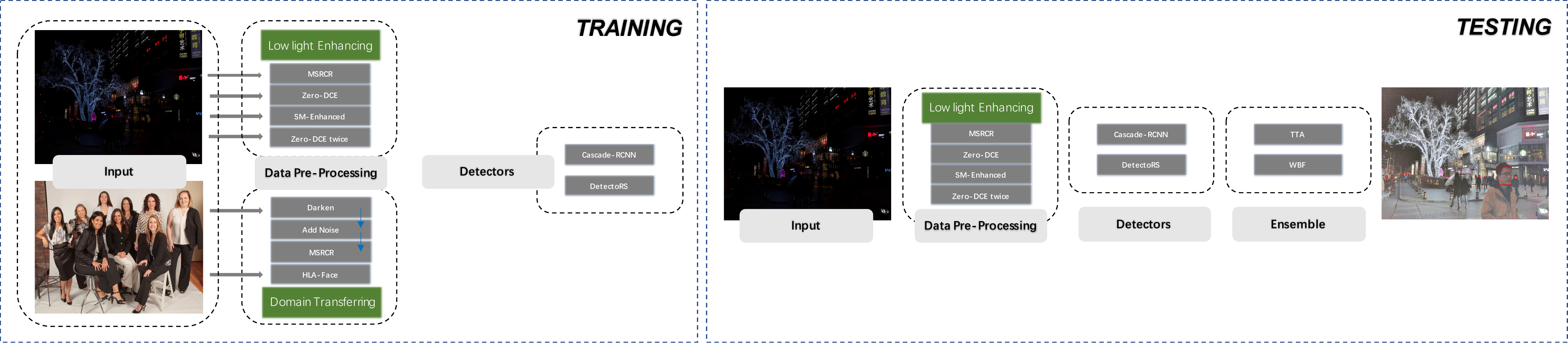}
        % \vspace{-0.3em}
        \caption{Whole Frameworks}
        \label{fig:fig2}
\vspace{-1em}
\end{figure*}
%%%%%%%%%%%%%%%%%%%%%%%

\subsection{Low Light Image Enhancement}

\begin{figure}[h] 
        \centering
        \includegraphics[width=1.0 \linewidth]{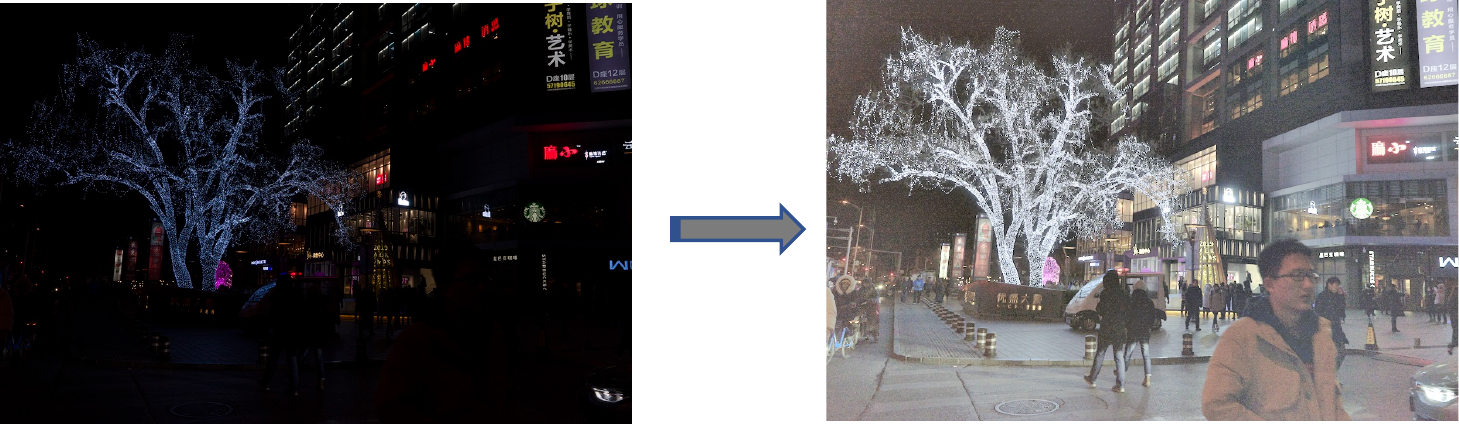}
        % \vspace{-0.1em}
        \caption{MSRCR method}
        \label{fig:fig3}
\vspace{-1em}
\end{figure}

To enhance the dark illumination of low light images, we employ the MSRCR \cite{597272} to it, which achieves simultaneous dynamic range compression/color consistency/lightness rendition. The enhanced image is shown in Figure  \ref{fig:fig3}. Besides, another data-driven brightness restoration method \cite{9157813} is also used, which formulating light enhancement as a task of image-specific curve estimation with a deep network. The brightness restoration result is shown in Figure \ref{fig:fig5}.

\begin{figure}[h] 
        \centering
        \includegraphics[width=1.0 \linewidth]{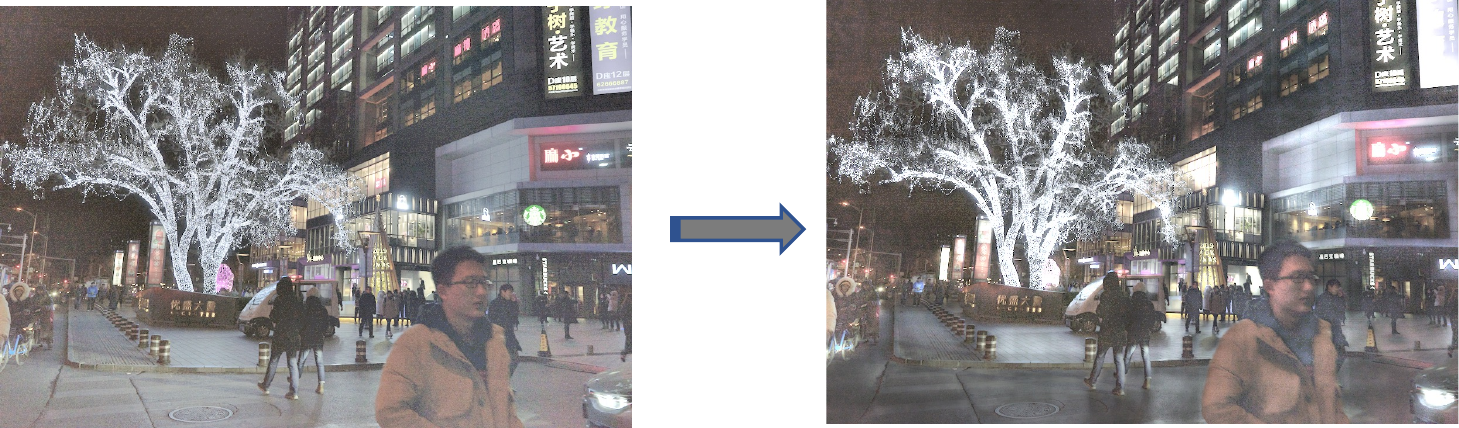}
        % \vspace{-0.3em}
        \caption{Saliency map enhanced}
        \label{fig:fig4}
\vspace{-1em}
\end{figure}

\begin{figure}[h] 
        \centering
        \includegraphics[width=1.0 \linewidth]{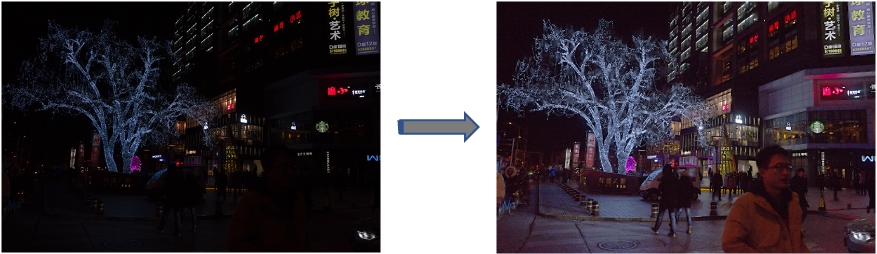}
        % \vspace{-0.3em}
        \caption{ZeroDCE method}
        \label{fig:fig5}
\vspace{-1em}
\end{figure}

Additionally, the saliency map $R_{saliency}$ \cite{5963689} extracted from the enhanced low light images $R_{msrcr}$, has been fused on $R_{msrcr}$ to suppress the false negative result, the fusion result $R_{saliency\_enhanced}$ is getting with:
\begin{equation}
R_{saliency\_enhanced} = \alpha * R_{saliency} + (1-\alpha )*R_{msrcr}
\end{equation}
where the $\alpha$ is set 0.3 in our work, and the result is shown in Figure \ref{fig:fig4}
%%

%-------------------------------------------------------------------------
\subsection{Normal Image Domain Transfer}

\begin{figure}[h] 
        \centering
        \includegraphics[width=1.0 \linewidth]{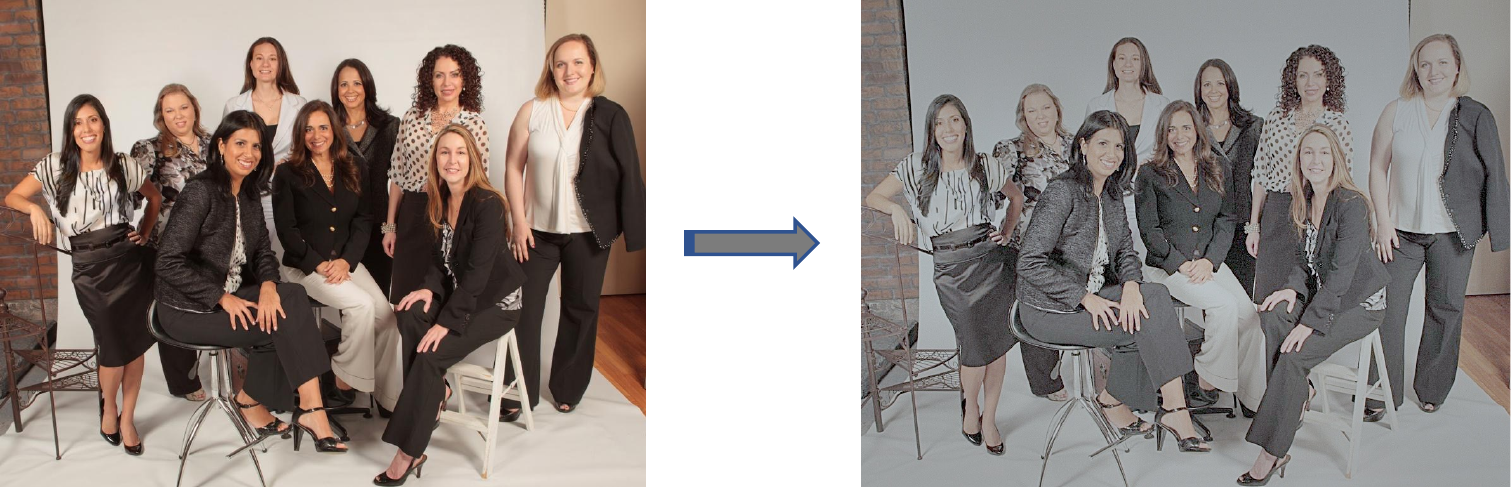}
        % \vspace{-0.3em}
        \caption{Traditional transfer method}
        \label{fig:fig6}
\vspace{-1em}
\end{figure}

\begin{figure}[h] 
        \centering
        \includegraphics[width=1.0 \linewidth]{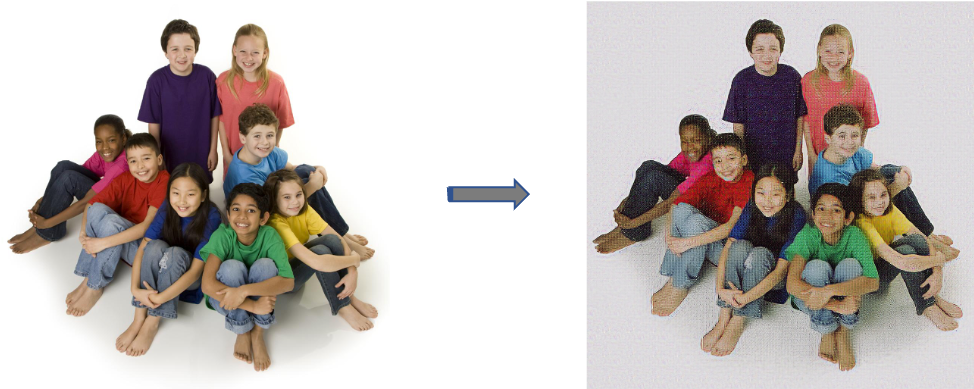}
        % \vspace{-0.3em}
        \caption{HLAFace method}
        \label{fig:fig7}
\vspace{-1em}
\end{figure}

% wpc改过
Different with using WIDERFACE and UFDD as pre-train sets, we merge them with pre-processed DARKFACE as a whole to build a more robust detector. Considering of the domain gap between pre-processed DARKFACE samples with the normal images (WIDERFACE, UFDD) that states in \cite{wang2021hlaface}, we transfer WIDERFACE and UFDD to a more closer domain of processed DARKFACE set firstly. There are two different ways to achieve it, the traditional one is darkening the normal images, adding noise, and then processing it with MSRCR \cite{597272}, the result is shown as Figure \ref{fig:fig6}. Another method like \cite{wang2021hlaface},  using the Pix2Pix network to synthesis noise, is shown in Figure \ref{fig:fig7}. 
Based on the above Low Light Enhancement and Domain Transfer methods, we can obtain training samples with more closer domain that consist of low light enhanced images and domain transferred normal images.
%

%%%%%%%%%%%%% figure 3 %%%%%%%%%%%%
%%%%%%%%%%%%%%%%%%%%%%%

%%%%%%%%%%%%% figure 3 %%%%%%%%%%%%
%%%%%%%%%%%%%%%%%%%%%%%

%%%%%%%%%%%%% figure 4 %%%%%%%%%%%%
%%%%%%%%%%%%%%%%%%%%%%%

%%%%%%%%%%%%% figure 5 %%%%%%%%%%%%
%%%%%%%%%%%%%%%%%%%%%%%

%%%%%%%%%%%%% figure 6 %%%%%%%%%%%%
%%%%%%%%%%%%%%%%%%%%%%%

%%%%%%%%%%%%% figure 8 %%%%%%%%%%%%
\begin{figure}[h] 
        \centering
        \includegraphics[width=1.0 \linewidth]{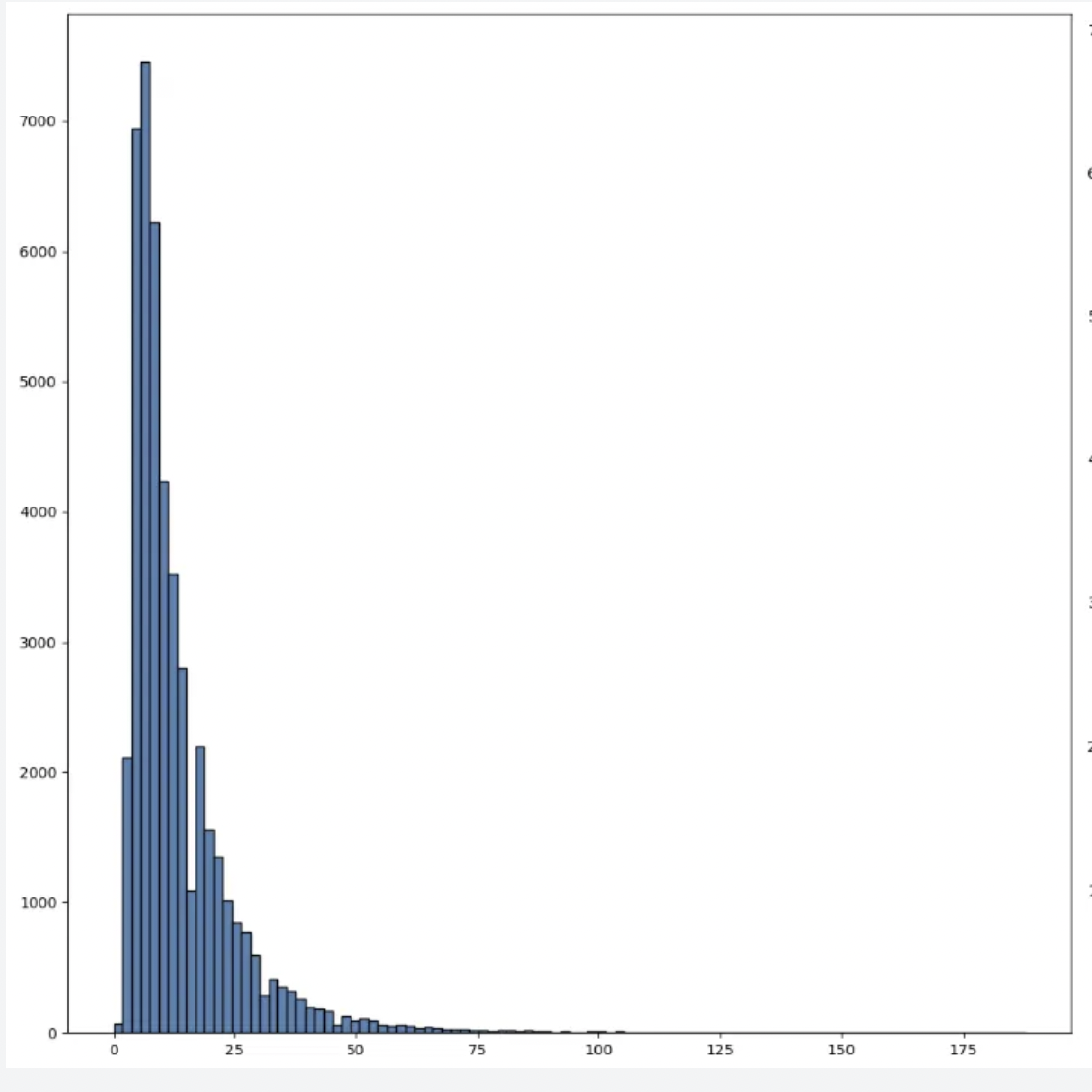}
        % \vspace{-0.3em}
        \caption{Distribution of face size(width): }
        \label{fig:fig8}
\vspace{-1em}
\end{figure}
%%%%%%%%%%%%%%%%%%%%%%%

\subsection{Detection methods}

We build a low light face detector based on the two-stage detection frameworks, including Cascade R-CNN \cite{8917599}, DetectoRS \cite{qiao2020detectors}. Using Cascade R-CNN as an example to describe the details, and the whole framework is shown in Figure \ref{fig:fig2}.

{\bfseries Dataset Split} Firstly, we split the DARKFACE set into several groups according to the number of faces in each image, and then randomly choose 10\% samples of each group as the validation part, and the rest 90\% data as the training part. We use the data augmentation method that described in section 2.1 to pre-process the DARKFACE samples. We also add WIDERFACE and UFDD dataset into our training set, which are pre-processed by the methods that described in section 2.2.

{\bfseries Training Strategy} We apply the multi-scale training tricks with resizing samples range from [2160,1440] to [4320,2880], and apply random crop with size [1000,800] on it. Additionally, we use the popular image augment tools \cite{Buslaev_2020} to process our training samples online, including random flip and random lightness, color jittering, and several filtering methods, etc. Using AdamW optimizer with 0.0001 as initial learning rate, linear decay in 27 and 33 epoch with total 36 epoch, and the weight decay is 0.05.

{\bfseries Model Refinement} Feature representation is always the key point to the object detection task. The backbone is very important to the feature representation ability of networks. Thus, we adopt two series powerful architecture Swin-Transformer \cite{liu2021swin} and ResNet \cite{xie2017aggregated} as our backbones. 
Besides, we also adapt PAFPN \cite{tan2020efficientdet} to replace FPN in Cascade R-CNN.
After analyzing the distribution of face size of the DARKFACE set, it is noticed that faces with small size is dominated which is illustrated in Figure \ref{fig:fig8}. Thus we set more small anchors to capture more tiny faces. We also add attention modules such as GCnet \cite{cao2019gcnet} in the backbone to obtain more powerful representations. RoI-align module is also adopt in order to predict more precise bounding boxes.

{\bfseries Model Ensemble} Finally, we trained Cascade R-CNN and DetectorRS with various backbones such as Swin-large, Swin-base, ResNet50 to obtain better diversity result of detectors. The performance of the models above that are trained on whole set can be found in the Table \ref{tab:tab1}. We use Weighted Boxes Fusion (WBF) \cite{solovyev2021weighted} and Test Time Augmentation(TTA) method for combining the predictions of our detectors, and the Soft-NMS\cite{8237855} is employed before the model ensemble process.

%%
%%GCnet$_{RoI-align}$
\begin{table}[h]
\centering
\caption{Validation results, where DetoRS means DetectoRS detection, CasR is Cascade R-CNN method, and GC$_{R}$ means GCnet is adopt in RoI-align module, Swin-b and Swin-l means Swin-base and Swin-large.}
\vspace{0.5em}
\begin{tabular}{|l|l|l|} 
\hline
Methods  &Setting         & mAP    \\ 
\hline
D-R      &DetoRS(Resnet50)       & 0.817  \\ 
\hline
D-RG    &DetoRS(Resnet50,GC$_{R}$)       & 0.819  \\ 
\hline
D-RGT    &DetoRS(Resnet50,GC$_{R}$,TTA)       & 0.821  \\ 
\hline
CR-S$_{b}$ &CasR(Swin-b)  & 0.81  \\ 
\hline
CR-S$_{b}$G &CasR(Swin-b,GC$_{R}$)  & 0.813  \\ 
\hline
CR-S$_{b}$GP &CasR(Swin-b,GC$_{R}$,PAFPN)  & 0.82  \\ 
\hline
CR-S$_{b}$GPT &CasR(Swin-b,GC$_{R}$,PAFPN,TTA)  & 0.823  \\ 
\hline
CR-S$_{l}$GP &CasR(Swin-l,GC$_{R}$,PAFPN) & 0.83   \\ 
\hline
CR-S$_{l}$GPT &CasR(Swin-l,GC$_{R}$,PAFPN,TTA) & 0.835   \\ 
\hline
WBF         &D-RGT,CR-S$_{b}$GPT,CR-S$_{l}$GPT         & 0.843  \\
\hline
\end{tabular}
\label{tab:tab1}
\end{table}

\section{Conclusion}

In our submission to the (Semi-) supervised Face detection in the low light condition in UG2$^{+}$ Challenge in CVPR 2021, we adopt two low light image enhancement methods to achieve brightness rendition. Besides, to obtain more training images, we transfer a lot of normal images (WIDERFACE and UFDD) to a closer domain with the brightness rendition images. Finally, we employ several powerful detectors to localize the bounding box of the face. In our future works, we will explore the end-to-end detection methods to process this work.

{\small
\bibliographystyle{ieee}
\bibliography{egbib}
}

\end{document}